\documentclass[10pt,twocolumn,letterpaper]{article}
\usepackage[pagenumbers]{cvpr}
\def\arxivpreprint{1}
\usepackage[OT1]{fontenc}
\usepackage{amsmath,amssymb,amsthm,booktabs,graphicx,xcolor}
\usepackage{array,colortbl,tabularx,multirow,placeins,etoolbox}
\definecolor{methodgray}{HTML}{EDEDED}
\definecolor{tablerowgray}{HTML}{F7F7F7}
\definecolor{tablerowblue}{HTML}{F2F7FB}
\definecolor{methodblue}{HTML}{DAEEF7}
\newcommand{\method}{SpikWM}
\newcommand{\methodfull}{Spiking Latent World Model}
\newcommand{\spikingbaseline}{Spiking-LeWM}
\newcommand{\sg}{\operatorname{sg}}

\newcommand{\ind}{\mathbf{1}}
\newcommand{\norm}[1]{\lVert#1\rVert}
\newcommand{\runin}[1]{\par\noindent\textbf{#1.}\enspace}
\newcommand{\supplementonly}[1]{#1}
\newtheoremstyle{compactprop}{5pt}{5pt}{\itshape}{}{\bfseries}{.}{.5em}{}
\theoremstyle{compactprop}

\AtBeginDocument{%
  \flushbottom
  \setlength{\parskip}{0pt plus 1pt}%
  \setlength{\abovedisplayskip}{5pt plus 2pt minus 2pt}%
  \setlength{\belowdisplayskip}{5pt plus 2pt minus 2pt}%
  \setlength{\abovedisplayshortskip}{3pt plus 2pt minus 1pt}%
  \setlength{\belowdisplayshortskip}{3pt plus 2pt minus 1pt}%
  \postdisplaypenalty=10000
  \setlength{\jot}{2pt}%
  \setlength{\textfloatsep}{12pt plus 2pt minus 2pt}%
  \setlength{\dbltextfloatsep}{12pt plus 2pt minus 2pt}%
  \setlength{\floatsep}{10pt plus 2pt minus 2pt}%
  \setlength{\dblfloatsep}{10pt plus 2pt minus 2pt}%
}
\definecolor{teal}{HTML}{347D80}
\definecolor{terra}{HTML}{C77358}
\definecolor{ink}{HTML}{263540}
\definecolor{pale}{HTML}{F0F5F5}

\usepackage{flushend}
\makeatletter
\newcommand{\cvpr@tagbalancedcolumns}{%
  \setbox\@leftcolumn\vbox to\ht\@leftcolumn{%
    \boxmaxdepth\@maxdepth
    \protected@write\@auxout{}{\string\def\string\@LN@column{1}}%
    \box\@leftcolumn}%
  \setbox\@outputbox\vbox to\ht\@outputbox{%
    \boxmaxdepth\@maxdepth
    \protected@write\@auxout{}{\string\def\string\@LN@column{2}}%
    \box\@outputbox}%
}
\apptocmd{\set@outputbox@with@footnote@and@float}
  {\cvpr@tagbalancedcolumns}
  {}{\PackageError{manuscript}{Could not patch flushend column tags}{}}
\makeatother

\renewcommand{\supplementonly}[1]{}
\definecolor{cvprblue}{rgb}{0.21,0.49,0.74}
\usepackage[breaklinks,colorlinks,allcolors=cvprblue]{hyperref}
\def\confName{CVPR}
\def\confYear{2027}
\title{SpikWM: Efficient End-to-End Spiking World Modeling for Visual Planning}
\author{Shuhan Ye \quad Hongbin Yu \quad Chong Wang \quad Hui Lu \quad Jiaming He\\
Yi Yu \quad Jiangbo Qian \quad Xudong Jiang\\[3pt]
{\tt\small shuhan006@e.ntu.edu.sg}}
\hypersetup{
  pdftitle={SpikWM: Efficient End-to-End Spiking World Modeling for Visual Planning},
  pdfauthor={Shuhan Ye, Hongbin Yu, Chong Wang, Hui Lu, Jiaming He, Yi Yu, Jiangbo Qian, Xudong Jiang}
}
\begin{document}
\maketitle
\begin{abstract}
Goal-conditioned planning with image observations and goals requires predictions that distinguish visual states and action consequences. Spiking neural networks offer sparse computation, but single-step spikes can obscure motion cues and merge actions with different physical outcomes. We propose the \methodfull{} (\method{}), an end-to-end joint-embedding predictive architecture (JEPA) with one simulation step per image and one neural update per latent position. Spiking Regional Encoding retains spatial cues through regional pooling. Bi-Spike Action Encoding combines positive and negative binary responses, while Causal Membrane Prediction carries subthreshold evidence across the latent history with learned group-wise time constants. The encoder and predictor are trained jointly on reward-free image--action trajectories using next-embedding prediction and SIGReg. With a common planner, \method{} achieves mean success rates of $88.67\%$, $97.33\%$, $99.33\%$, and $86.67\%$ on Push-T, Two-Room, Reacher, and OGBench-Cube, respectively, outperforming two spiking baselines. Push-T ablations support each module, and controlled replays examine responses to appearance and physical state changes. Under a matched planning budget, \method{} reduces estimated arithmetic energy by $75.6\%$ relative to LeWM.
\ifdefined\arxivpreprint
Code will be released at \url{https://github.com/ShawnYE618/SpikWM}.
\fi
\end{abstract}

\section{Introduction}
\label{sec:introduction}
Goal-conditioned planning from images requires choosing actions that lead to a desired state specified by a goal image. Reaching a target object pose requires anticipating how successive contacts change its position and orientation. World models (WMs) learn these action-dependent dynamics from experience, enabling agents to evaluate candidate plans before acting~\cite{hafner2019planet,hafner2020dreamcontrollearningbehaviors,nguyen2021tpc}. To make this evaluation tractable, latent WMs predict in a compact representation space without reconstructing future images~\cite{dinowm2025,sobal2025pldm,maes2026lewm}. Action-conditioned joint-embedding predictive architectures (JEPAs) learn these representations and dynamics from image--action trajectories by predicting future observation embeddings conditioned on preceding observations and actions~\cite{assran2023ijepa,maes2026lewm}. The planner then rolls out candidate action sequences and ranks their predicted endpoints by distance to the goal embedding, reusing the same dynamics across visual goals. However, this search requires repeated predictions across many candidates and rollout steps, making inference efficiency critical to planning latency and energy consumption.

\begin{figure}[t]
\centering
\includegraphics[width=\columnwidth]{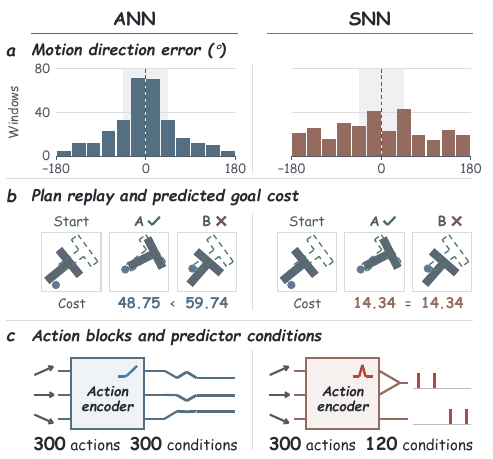}
\vspace{-6mm}
\caption{\textbf{Lost motion and action distinctions on Push-T.} LeWM (ANN) and \spikingbaseline{} (SNN).
(a) Motion direction readout errors on $305$ shared moving windows.
(b) Replayed endpoints and predicted goal costs. Dashed outlines mark the goal.
(c) Distinct predictor conditions for one pool of action blocks.}
\label{fig:prediction_overview}
\end{figure}

Spiking neural networks (SNNs) communicate through sparse, asynchronous spikes and integrate successive inputs through membrane dynamics~\cite{fang2021plif}. On neuromorphic hardware, this event-driven operation is energy-efficient~\cite{eshraghian2023snn}, making SNNs a potentially efficient alternative to dense networks for repeated prediction during planning. SNNs already perform competitively on visual tasks, including image classification, detection and segmentation~\cite{zhou2023spikformer,yao2023sdt,yao2024sdtv2}, as well as video recognition~\cite{zou2025spikevideoformer,ye2026fire}. Spiking-WM extends the use of SNNs to visual control, following Dreamer in using imagined rollouts to train a policy that maximizes task reward~\cite{sun2025spiking}. However, using spiking latent world models for goal-conditioned planning remains largely unexplored. In this setting, the planner uses the model at test time to rank candidate action sequences by how closely their predicted outcomes match a supplied goal image. Reliable ranking therefore requires a representation that retains goal-relevant details such as object pose and a predictor that distinguishes the effects of different actions over multiple steps.

Visual SNNs commonly repeat each image over $T$ simulation steps to accumulate neural responses~\cite{zhou2023spikformer,xu2024feelsnn}. Extending such multi-step computation to candidate rollouts increases planning cost. We therefore target one simulation step per image and one neural update per latent position. To examine the challenges of this setting, we replace selected continuous activations in LeWM with single-update spiking neurons, yielding \spikingbaseline{}. On Push-T, this direct substitution weakens motion-direction readout from visual latents (Fig.~\ref{fig:prediction_overview}a) and maps $300$ distinct action blocks to only $120$ predictor conditions (Fig.~\ref{fig:prediction_overview}c). The latter has a concrete planning consequence: two plans differing only in their first action block receive identical forecasts and goal costs, although only one reaches the goal (Fig.~\ref{fig:prediction_overview}b). On fixed action sequences, median fifth-block prediction error is higher with predicted history than with observed history (Sec.~\ref{sec:theory}). These findings highlight the central challenge: \emph{reducing neural updates while retaining the information needed to distinguish and rank candidate plans}.

To address this challenge, we propose \methodfull{} (\method{}), an end-to-end JEPA-style model for goal-conditioned planning. We design visual encoding, action conditioning, and latent prediction to preserve planning-relevant information. Since global pooling can obscure spatial changes, we introduce Spiking Regional Encoding (SRE, $\mathcal{E}$), which assigns channel groups to image regions and pools features within each region. Action conditioning faces a different bottleneck: positive-only thresholds can map distinct action currents to the same silent response. Bi-Spike Action Encoding (BAE, $\mathcal{A}$) therefore introduces complementary positive and negative spike streams, combining them through learned readouts to retain these distinctions. Within the predictor, evidence that is insufficient to trigger a spike at one position may still inform later predictions. Causal Membrane Prediction (CMP, $\mathcal{D}$) preserves this evidence through membrane integration with learned group-wise time constants, alongside causal spike-driven attention. The resulting model processes each image in one simulation step and performs one neural update per context position. We evaluate planning success on four tasks and examine the proposed designs through ablations and controlled replays.

Our main contributions are summarized as:
\begin{itemize}
\item We propose \method{}, to our knowledge the first spiking latent world model for goal-conditioned visual planning. Its efficient architecture uses one simulation step per image and one neural update per context position.
\item We identify weakened motion cues, action-condition collisions, and errors from imagined histories. Based on these findings, we design SRE, BAE, and CMP.
\item \method{} surpasses both spiking baselines on four tasks and many ANN baselines on multiple tasks. Estimated Push-T arithmetic energy is $75.6\%$ below LeWM's. Ablations and visualizations reveal distinct responses to appearance and physical state changes.
\end{itemize}

\section{Related Work and Preliminaries}
\label{sec:related}
\subsection{Latent World Models}
\label{sec:formulation}
Latent world models learn environment dynamics in a representation space. PlaNet uses observation and reward prediction for planning, while Dreamer learns policies from imagined latent trajectories~\cite{hafner2019planet,hafner2020dreamcontrollearningbehaviors}. JEPAs instead predict the representation of a target from an encoded context~\cite{assran2023ijepa}. Action-conditioned models extend this formulation to predicting the consequences of commands. DINO-WM and V-JEPA~2-AC learn dynamics on pretrained visual features~\cite{dinowm2025,assran2025vjepa2}. PLDM and LeWM jointly learn visual representations and dynamics from offline image--action trajectories~\cite{sobal2025pldm,maes2026lewm}. LeWM uses next-embedding prediction and SIGReg to train without reconstruction, stop-gradient or a target encoder. ALeWM further organizes predictive information into compact latent prefixes~\cite{achituve2026adaptivelatentcapacityworld}. We use LeWM as the ANN baseline and study the representation and temporal computation needed to construct a spiking model.

\vspace{1mm}\runin{Goal-conditioned planning}
Let $o_t$ be an observation image and $a_t$ the action block leading to $o_{t+1}$. Encoder $f_\theta$ and predictor $g_\phi$ share a $d$-dimensional latent space:
\begin{equation}\small
 z_t=f_\theta(o_t),\qquad
 \hat z_{t+1}=g_\phi(z_{\ell:t},a_{\ell:t}).
 \label{eq:latent_prediction}
\end{equation}
The context starts at $\ell=\max(1,t-L+1)$ and contains up to $L$ positions. 
The \method{} encoder and predictor (\Cref{sec:method_overview}) are jointly trained to match $\hat z_{t+1}$ to $z_{t+1}$ while regularizing the latents~\cite{maes2026lewm,balestriero2025lejepa}. At planning time, rolling out an action sequence $\mathbf a=(a_t,\ldots,a_{t+K-1})$ yields the predicted endpoint $\hat z_{t+K}(\mathbf a)$. For goal image $o_g$, the terminal cost is
\begin{equation}\small
 \hat J(\mathbf a;z_g)=\norm{\hat z_{t+K}(\mathbf a)-z_g}_2^2,
 \qquad z_g=f_\theta(o_g).
 \label{eq:planning_cost}
\end{equation}
CEM samples and refines candidate sequences under this cost~\cite{rubinstein1999cem}. The quality of the action therefore depends on both the visual representation and its predicted evolution.

\subsection{Spiking Neural Networks}
SNNs combine binary events with stateful neuronal dynamics. Surrogate-gradient learning enables optimization through hard firing decisions~\cite{neftci2019sg}, while learnable membrane time constants adapt temporal integration~\cite{fang2021plif}. Spiking Transformers extend these mechanisms to image recognition and dense prediction through spike-driven attention~\cite{yao2023sdt,yao2024sdtv2,yao2025sdtv3}. Video models further exploit neuronal state across successive frames for sequence processing~\cite{yu2024svformerdirecttrainingspiking,zou2025spikevideoformer}. Alongside these, population-coded control and multi-threshold neurons enrich input representations beyond a single binary threshold~\cite{tang2021popsan,fan2025msf,zhang2026hmt}. In recognition, encoding and leak can improve corruption tolerance, whereas suppressing high-frequency features can remove useful image detail~\cite{xu2024feelsnn,fang2025highfrequency}. For planning, the key question is whether such information loss changes the ranking of physical outcomes.

\vspace{1mm}\runin{Membrane dynamics}
A leaky integrate-and-fire (LIF)~\cite{fang2021plif} neuron retains part of its membrane potential between steps and resets after firing. With membrane time constant $\tau$ in step units and input increment $x_j$, the zero-reset update is
\begin{equation}\small
 \begin{aligned}
 u_j = \bigl(1-\tfrac{1}{\tau}\bigr)v_{j-1}+x_j,
 ~s_j = \ind[u_j\geq V_{\mathrm{th}}],~v_j=(1-s_j)u_j.
 \end{aligned}
 \label{eq:lif_prelim}
\end{equation}
Here $u_j$ and $v_j$ are the membrane potentials before firing and after reset, and $V_{\mathrm{th}}$ is the firing threshold. An integrate-and-fire (IF) neuron instead retains the entire previous membrane state without leakage. With one update from $v_0=0$, leakage has no effect; LIF and IF produce identical spikes for the same input increment and threshold. For frame sequences, the step index $j$ can follow observation order~\cite{yu2024svformerdirecttrainingspiking}. Our predictor carries membrane state across latent positions despite one update per position, retaining group-wise leakage with a subtractive reset.

\vspace{1mm}\runin{Spiking Prediction \& WMs}
Spiking prediction has been explored for video representations, dynamic graphs and images~\cite{dong2026prednext,zhang2026sgjepa,gleyo2026spikingjepa}. Earlier spiking planners operate on explicit task states or robot configurations~\cite{rueckert2016planning,tanneberg2016humanoids}, and generative spiking models support discrete visuomotor prediction~\cite{basanisi2020planning}. Spiking dynamics have also been used for learning from imagined experience and reward-guided control~\cite{capone2024dreaming,blakowski2025dreaming,buccapatnam2026spiking}. Spiking-WM combines a spiking visual encoder and state-space model with reconstruction, reward prediction and policy learning~\cite{sun2025spiking}. SpikeWorld corrects actions of a frozen policy under actuator changes~\cite{spikeworld2026}. We jointly learn visual representations and action-conditioned dynamics from reward-free trajectories, then use the predicted representations directly to score candidate plans for an image goal.

\section{Analysis and Motivation}
\label{sec:theory}
\begin{figure*}[t]
\centering
\includegraphics[trim=11bp 0bp 13bp 0bp,clip,width=\textwidth]{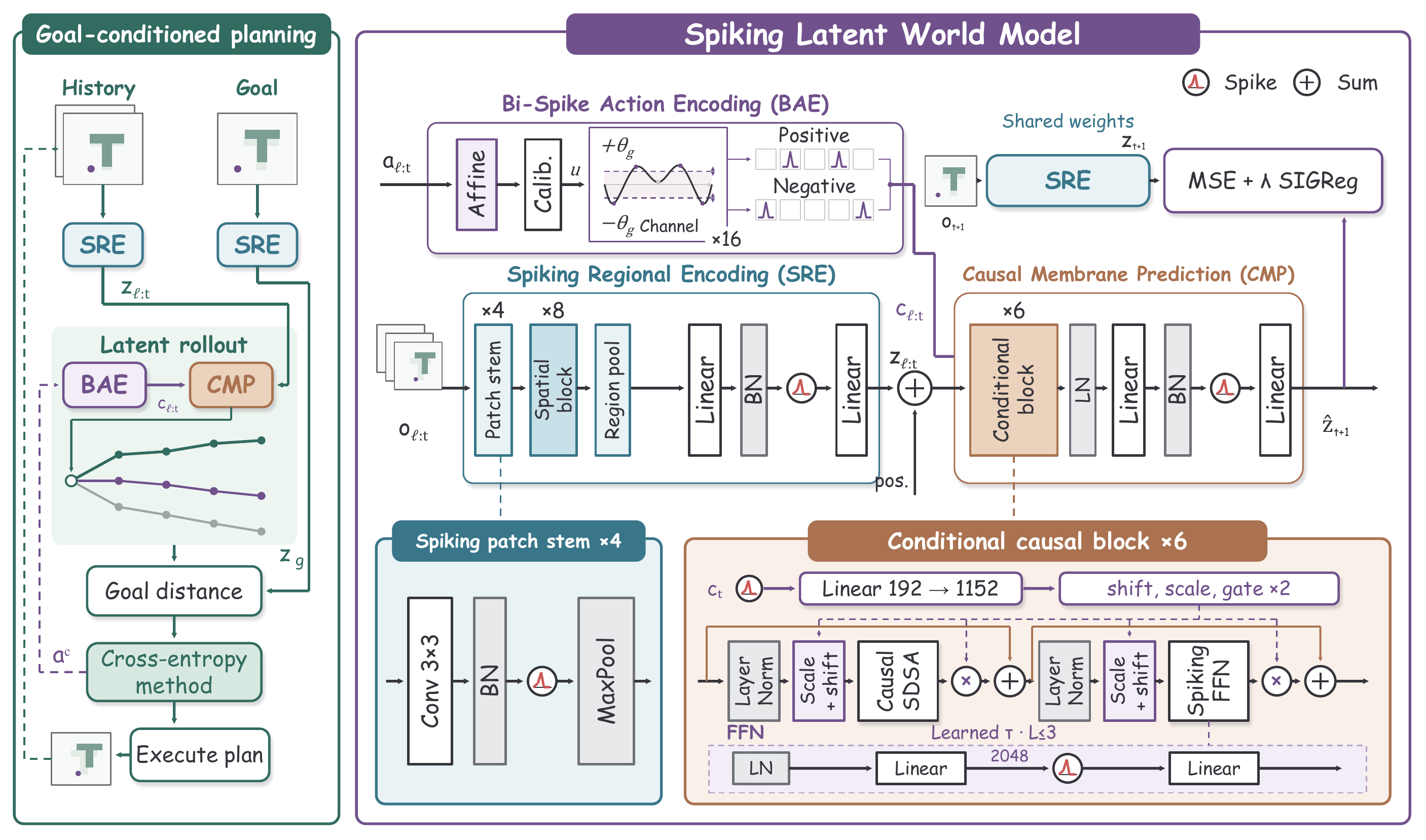}
\vspace{-6mm}
\caption{\textbf{\method{} architecture and planning.}
Spiking Regional Encoding (SRE) combines spatial spike processing, regional pooling and an output projection. Bi-Spike Action Encoding (BAE) uses learned group thresholds and combines two binary streams through separate maps. Causal Membrane Prediction (CMP) includes conditional causal blocks, final normalization and an output projection. Within each block, LayerNorm precedes action-conditioned scaling and shifting. Shared SRE branches learn from next-latent prediction and SIGReg. The cross-entropy method replans using terminal goal distance.}
\label{fig:n22_architecture}
\end{figure*}

Goal-conditioned planning ranks action sequences by the distance between predicted endpoints and the goal. This ranking can fail when distinct physical outcomes receive identical predictions, or when prediction errors reverse their ordering. We examine how visual encoding, action conditioning, and prediction history cause failures and motivate our design.

\vspace{1mm}\runin{Visual distinctions under small displacements}
Global average pooling discards where an activation occurs. Suppose a small displacement moves an activation of magnitude $\eta>0$ from outside into channel $c$'s assigned region $\Omega_r$, with other activations fixed. The global and regional averages over $\Omega$ and $\Omega_r$ then change by $\frac{\eta-\eta}{|\Omega|}=0,~[\Delta e_t]_c=\frac{\eta}{|\Omega_r|}$.
Regional pooling can thus retain shifts across region boundaries that global pooling suppresses. The linear motion probe in Fig.~\ref{fig:prediction_overview}a also shows less accurate direction readout from \spikingbaseline{}'s latents than from LeWM's. Together, these observations motivate preserving spatial cues through regional pooling in SRE $\mathcal{E}$ (\cref{sec:visual_encoding}).

\vspace{1mm}\runin{When distinct actions give the same prediction}
Action-conditioned world models predict subsequent states from visual history and candidate actions~\cite{maes2026lewm}. Let $\mathcal{C}(a)$ denote all action-dependent inputs received by the predictor for action block $a$. Under the same visual history and initial predictor state, $\mathcal{C}(a)=\mathcal{C}(b)$ implies identical one-step predictions for a deterministic predictor. If this equality holds at every rollout step, the two sequences produce identical predicted terminal states and goal costs.
For a fixed encoder $f_\theta$, let $z^{(j)}=f_\theta(o_{t+1}^{(j)})$ denote the successor embedding associated with action block $j\in\{a,b\}$. When both actions share a prediction, the minimum per-coordinate MSE is
\vspace{-1mm}
\begin{equation}\small
 \min_{\hat z}\frac{1}{2d}\sum_{j\in\{a,b\}}\norm{\hat z-z^{(j)}}_2^2
 =\frac{1}{4d}\norm{z^{(a)}-z^{(b)}}_2^2.
 \label{eq:shared_prediction_minimum}
\end{equation}
The minimum is attained at $\tfrac{(z^{(a)}+z^{(b)})}{2}$ and is positive when $z^{(a)}\ne z^{(b)}$. In the \spikingbaseline{} replay in Fig.~\ref{fig:prediction_overview}b, two action sequences share their start, goal, and final twenty commands but differ in the first five. Despite identical action conditions and predictions across five blocks, only one sequence reaches the goal, and their final object centers differ by $60.8$ simulator pixels. Positive-only thresholding can induce such collisions by mapping strongly negative and near-zero currents to the same zero response, motivating the additional negative-current spike stream in BAE $\mathcal{A}$ (\cref{sec:population_placement}) to preserve action distinctions for visual feature modulation (\cref{eq:conditional_block}). Prediction errors can also reverse the ranking of sequences with distinct predicted outcomes.

\vspace{1mm}\runin{Effect of predicted history on rollout error}
For a fixed action sequence, let $Z_t=z_{\ell:t}$ denote the encoded replay history. The rollout history $\widetilde Z_t$ replaces each unavailable latent $z_j$ with its prediction $\hat z_j$. Writing $g_t(Z)=g_\phi(Z,a_{\ell:t})$ for the predictor in \cref{eq:latent_prediction}, the rollout error decomposes as
\begin{equation}\small
 \begin{gathered}
 \!\!\!\varepsilon_{t+1}\!=\!g_t(\widetilde Z_t)\!-\!z_{t+1}
 \!=\!\bigl[g_t(\widetilde Z_t)\!-\!g_t(Z_t)\bigr]
   \!+\!\bigl[g_t(Z_t)\!-\!z_{t+1}\bigr].
 \end{gathered}
 \label{eq:history_error_decomposition}
\end{equation}
The first term captures the change in prediction induced by replacing replay history with predicted history. The second is the prediction error under replay history. On $24$ fixed-action plans, using replay history reduces \spikingbaseline{}'s fifth-block median latent $\ell_2$ error from $0.689$ to $0.252$ after normalization by the start-to-goal distance $\norm{z_g-z_{\mathrm{start}}}_2$.
These results highlight the importance of temporal context. To retain temporal evidence under single-step computation, CMP $\mathcal{D}$ carries feed-forward membrane states across positions within the context window. Subthreshold contributions can then decay and combine with later inputs to influence subsequent firing decisions (\cref{eq:real_history_lif,eq:causal_membrane_history}).

\section{\methodfull{} (\method{})}
\label{sec:method}

\subsection{Overview}
\label{sec:method_overview}
To retain visual and action distinctions during latent rollouts revealed in \cref{sec:theory}, we propose the \method{} for goal-conditioned planning. As shown in Fig.~\ref{fig:n22_architecture}, Spiking Regional Encoding (SRE, $\mathcal{E}$) maps each frame to a latent through regional pooling and an output projection in one simulation step (\cref{sec:visual_encoding}). Bi-Spike Action Encoding (BAE, $\mathcal{A}$) maps each action block to a conditioning vector (\cref{sec:population_placement}). Causal Membrane Prediction (CMP, $\mathcal{D}$) predicts successor latents from visual and action histories with one neural update per context position (\cref{sec:causal_dynamics}).

At each position $t$, the inputs are an RGB frame $o_t\!\in\!\mathbb R^{3\times224\times224}$ and the corresponding action block $a_t$. SRE produces $z_t\!=\!\mathcal{E}(o_t)$, so $f_\theta\!=\!\mathcal{E}$ in \cref{eq:latent_prediction}. The action condition is $c_t\!=\!\mathcal{A}(a_t)$. Given the visual history $z_{\ell:t}$ and action conditions $c_{\ell:t}$, CMP produces the successor latent
\begin{equation}\small
 \hat z_{t+1}=\bigl[\mathcal{D}(z_{\ell:t},c_{\ell:t})\bigr]_t.
 \label{eq:ead_chain}
\end{equation}
The subscript $t$ selects the output at the last context position.

\vspace{1mm}\runin{Training and planning}
\label{sec:learning_planning}
We train on four-frame windows with $L=3$ observation--action pairs. At each position, the model predicts the next latent from observed history and matches it to the next encoded frame. For a batch of $B$ windows, the objective is
\begin{equation}\small
 \mathcal{L}=\frac{1}{BLd}\sum_{n,t}\norm{\hat z_{n,t+1}-z_{n,t+1}}_2^2
 +\lambda\,\mathcal{L}_{\mathrm{SIGReg}}.
 \label{eq:training_loss}
\end{equation}
The sum ranges over windows $n$ and their $L$ prediction positions $t$. $\mathcal{L}_{\mathrm{SIGReg}}$ regularizes the embeddings at each observation position with weight $\lambda=0.09$~\cite{maes2026lewm,balestriero2025lejepa}. Gradients reach the shared SRE through the prediction and target branches, with surrogate derivatives at hard thresholds~\cite{neftci2019sg}.
At planning time, we encode the goal as $z_g=\mathcal{E}(o_g)$. Each candidate rollout appends the forecast from \cref{eq:ead_chain} and keeps the most recent $L$ positions. CEM refines candidate action sequences using the terminal goal cost in \cref{eq:planning_cost}. The agent executes the five-block plan ($25$ commands) before replanning from new observations.

\subsection{Spiking Regional Encoding (SRE)}
\label{sec:visual_encoding}
The encoder $\mathcal{E}$ resets before each frame. We use $\operatorname{Conv}$ for a padded $3\times3$ convolution with stride one, $\operatorname{BN}$ for batch normalization, and $\operatorname{MaxPool}_2$ for max pooling with stride two. The operator $\operatorname{LIF}$ returns the spike from one zero-state update of \cref{eq:lif_prelim}. Starting from $x_0=o_t$, the four stem stages ($k=1,\ldots,4$) and the positional residual compute
\begin{equation}\small
 \begin{gathered}
 x_k=\operatorname{MaxPool}_2\!\bigl(\operatorname{LIF}(\operatorname{BN}(\operatorname{Conv}(x_{k-1})))\bigr),\\
 x=x_4+\operatorname{LIF}(\operatorname{BN}(\operatorname{Conv}(x_4))).
 \end{gathered}
 \label{eq:sre_stem}
\end{equation}
The convolution and batch normalization layers have separate parameters at each stage and in the positional branch.

Eight spatial blocks then process the grid. Each applies spike-driven self-attention (SDSA)~\cite{yao2023sdt} across the positions in one frame, followed by a spiking feed-forward network (FFN). The two residual updates are
\begin{equation}\small
 y=x+\operatorname{SDSA}(x), \qquad x'=y+\operatorname{FFN}(y).
 \label{eq:sre_spatial_block}
\end{equation}
The output of the eighth block is $F_t\in\mathbb R^{192\times14\times14}$.

Global average pooling averages each channel over the entire grid. Our regional average pooling assigns a disjoint group $\mathcal{C}_r$ of $48$ channels to each of the four $7\times7$ quadrants $\Omega_r$. For channel $c\in\mathcal{C}_r$, the mean uses only positions in its assigned quadrant,
\begin{equation}\small
 [e_t]_c=\frac{1}{|\Omega_r|}\sum_{p\in\Omega_r}[F_t]_{c,p}.
 \label{eq:spatial_readout}
\end{equation}
We concatenate the four groups into $e_t$. An internal spiking MLP maps this vector to the SRE output $z_t$. It uses two affine layers with batch normalization and a calibrated binary spike activation between them. CMP then uses $z_t$ with the condition $c_t$ from BAE.

\begin{table*}[t]
\centering
\caption{\textbf{Goal-conditioned planning success.} SNN results are mean $\pm$ sample SD across three evaluation seeds. $\dagger$ denotes Anchor-WM's five-seed results~\cite{wang2026anchorwm}. $\ddagger$ marks our Spiking-WM adaptation. Updates count ANN passes or SNN steps per latent position. Parameters cover the deployed Push-T modules. Bold marks the highest mean within each group.}
\vspace{-2mm}
\label{tab:planning_results}
\small
\definecolor{spikecross}{HTML}{B3261E}
\definecolor{spikecheck}{HTML}{167343}
\newcommand{\nonspiking}{\textcolor{spikecross}{$\times$}}
\newcommand{\spikingyes}{\textcolor{spikecheck}{$\checkmark$}}
\newcommand{\anchorretest}{\textsuperscript{\ensuremath{\dagger}}}
\newcommand{\offlineadapt}{\textsuperscript{\ensuremath{\ddagger}}}
\setlength{\tabcolsep}{3pt}
\renewcommand{\arraystretch}{1.10}
\begin{tabularx}{\textwidth}{@{}lcccr*{4}{>{\centering\arraybackslash}X}@{}}
\toprule
& & & & & \multicolumn{4}{c}{Success (\%) $\uparrow$}\\
\cmidrule(l){6-9}
Method & Spiking & Updates & Features & Params (M) & Push-T & Two-Room & Reacher & \mbox{OGBench-Cube}\\
\midrule
PLDM\anchorretest~\cite{sobal2025pldm} & \nonspiking & 1 & Learned & 18.03 & $71.8_{\pm6.1}$ & $93.0_{\pm2.0}$ & $69.5_{\pm5.8}$ & $51.2_{\pm1.6}$\\
DINO-WM\anchorretest~\cite{dinowm2025} & \nonspiking & 1 & DINOv2 & 41.75 & $65.2_{\pm6.6}$ & $99.6_{\pm0.9}$ & $67.2_{\pm1.8}$ & $74.4_{\pm2.6}$\\
Sub-JEPA~\cite{zhao2026subjepa} & \nonspiking & 1 & Learned & 18.03 & $89.00_{\pm5.33}$ & $95.00_{\pm2.76}$ & $84.00_{\pm4.00}$ & $76.33_{\pm5.99}$\\
Fast-LeWM~\cite{gao2026fastlewm} & \nonspiking & 1 & Learned & 17.91 & $91.7_{\pm4.1}$ & $95.4_{\pm3.2}$ & $84.0_{\pm5.9}$ & $71.7_{\pm8.4}$\\
$\Sigma$-LeWM~\cite{zu2026sigmalewm} & \nonspiking & 1 & Learned & 18.03 & $\mathbf{94.5}_{\pm3.0}$ & $92.4_{\pm4.4}$ & $86.0_{\pm4.8}$ & $82.6_{\pm5.0}$\\
RC-aux~\cite{li2026rcaux} & \nonspiking & 1 & Learned & 18.71 & $90.8_{\pm3.3}$ & $98.0_{\pm1.4}$ & $87.2_{\pm6.4}$ & $76.0_{\pm7.5}$\\
TD-JEPA~\cite{bai2026temporaldistancejepa} & \nonspiking & 1 & Learned & 18.03 & $86.0_{\pm4.2}$ & $\mathbf{100.0}_{\pm0.0}$ & $\mathbf{97.0}_{\pm2.4}$ & $82.2_{\pm2.9}$\\
Qantara~\cite{rakhimov2026qantara} & \nonspiking & 4 & Learned & 21.54 & $90.1_{\pm1.1}$ & $\mathbf{100.0}_{\pm0.0}$ & $80.9_{\pm1.8}$ & $\mathbf{93.7}_{\pm0.7}$\\
LeWM\anchorretest~\cite{maes2026lewm} & \nonspiking & 1 & Learned & 18.03 & $90.4_{\pm4.3}$ & $86.1_{\pm1.8}$ & $78.4_{\pm5.7}$ & $64.8_{\pm2.1}$\\
\midrule
Spiking-WM\offlineadapt~\cite{sun2025spiking} & \spikingyes & 5 & Learned & 19.95 & $37.33_{\pm6.11}$ & $92.67_{\pm1.15}$ & $70.67_{\pm7.57}$ & $52.00_{\pm2.00}$\\
\spikingbaseline{} & \spikingyes & 1 & Learned & 18.03 & $53.33_{\pm1.15}$ & $88.00_{\pm2.00}$ & $70.67_{\pm6.11}$ & $72.67_{\pm7.57}$\\
\rowcolor{methodgray}
\textbf{\method{} (ours)} & \spikingyes & 1 & Learned & 16.81 & $\mathbf{88.67}_{\pm1.15}$ & $\mathbf{97.33}_{\pm3.06}$ & $\mathbf{99.33}_{\pm1.15}$ & $\mathbf{86.67}_{\pm1.15}$\\
\bottomrule
\end{tabularx}
\end{table*}

\subsection{Bi-Spike Action Encoding (BAE)}
\label{sec:population_placement}
BAE $\mathcal{A}$ uses two binary spike streams with learnable thresholds to distinguish actions within one update.

\vspace{1mm}\runin{Two binary spike streams}
We standardize the five commands in $a_t$ using training-set statistics and concatenate them as $\bar a_t$. A learned affine map projects this vector to $768$ channels. Fixed calibration vectors $\mu$ and $\sigma>0$ give the current
\begin{equation}\small
 u_t=\frac{W_a\bar a_t+b_a-\mu}{2\sigma}.
 \label{eq:action_calibration}
\end{equation}
Division and the threshold comparisons below act elementwise. We divide the channels into $16$ groups of $48$. Each group learns a threshold $\theta_g=\tfrac12(1+\tanh\beta_g)$ through the prediction loss, with $\beta_g=0$ initially. Repeating each group's threshold across its channels gives the vector $\theta$, initially $0.5$ in every coordinate. The two spike streams are
\begin{equation}\small
 p_t=\ind[u_t\geq\theta], \qquad
 n_t=\ind[u_t\leq-\theta].
 \label{eq:population_mapping}
\end{equation}
Both $p_t$ and $n_t$ contain only zeros and ones. They respond to the positive and negative sides of the calibrated current, respectively. At each channel, the possible output pairs are $(1,0)$, $(0,1)$ and $(0,0)$.

For the code comparison, we hold the learned projection and thresholds fixed. The joint code $Q=(p_t,n_t)$ contains the first stream's code $Q_0=p_t$, and the second stream can distinguish actions with identical $p_t$. At a fixed context, suppose $Q$ splits one such class into two with conditional masses $\pi_1,\pi_2>0$ and mean successor latents $\bar z_1,\bar z_2$. For a fixed target encoder, let $\Delta\mathcal{R}$ be the decrease in that class's contribution to the minimum per-coordinate prediction risk. Then we have
 $\Delta\mathcal{R}=
 \frac{\pi_1\pi_2}{d(\pi_1+\pi_2)}\norm{\bar z_1-\bar z_2}_2^2$.
With just two equally likely targets, one per class, this equals the shared-prediction error in \cref{eq:shared_prediction_minimum}.

\vspace{1mm}\runin{Combining the spike streams}
Separate learned maps $W,V\in\mathbb R^{d\times768}$ combine the two streams into the action condition:
 $c_t=Wp_t+Vn_t+b$.
We initialize $V=0$, so initially $c_t=Wp_t+b$. At a single training position, a nonzero $n_t$ and a nonzero gradient at $c_t$ give a nonzero gradient for $V$, even at initialization. The gradient reaching the second stream is initially zero and depends on both $V$ and the downstream signal thereafter.

\vspace{1mm}\runin{From spikes to modulation}
For two action blocks $a$ and $a'$ with identical first-stream outputs, write $\Delta c\!=\!c(a)\!-\!c(a')$ and $\Delta n\!=\!n(a)\!-\!n(a')$. Their condition difference is
 $\Delta c\!=\!V\Delta n$.
Every block in $\mathcal{D}$ computes $\xi_t\!=\!\ind[c_t\!\geq\!\theta_c]$ from zero membrane state, with fixed threshold $\theta_c\!=\!0.125$. At one coordinate, order the two condition values as $c\leq c'$. Their gate outputs differ exactly when
 $c\!<\!\theta_c\!\leq\! c'$.
The distinction reaches visual modulation if at least one block maps the two gate vectors to different modulation parameters (\cref{eq:conditional_block}).

\subsection{Causal Membrane Prediction (CMP)}
\label{sec:causal_dynamics}
$\mathcal{D}$ follows LeWM's six-block causal layout~\cite{maes2026lewm}. It adds learned position embeddings to the visual latents, and each block maps $\xi_t$ to shift, scale and gate vectors for its attention and feed-forward branches. Write the adaptive normalization as $M(x;b,\eta)=(1+\eta)\odot\operatorname{LN}(x)+b$. With position and block indices omitted, the updates are
\begin{equation}\small
 \begin{aligned}
 y  &= x+g_1\odot\operatorname{Attn}\bigl(M(x;b_1,\eta_1)\bigr),\\
 x' &= y+g_2\odot\operatorname{FFN}\bigl(M(y;b_2,\eta_2)\bigr).
 \end{aligned}
 \label{eq:conditional_block}
\end{equation}
A final LayerNorm and an internal spiking output MLP follow the six blocks. The MLP uses the same layer pattern as SRE's output projection, with separate parameters, and maps each context position to its successor latent.

\vspace{1mm}\runin{Causal spike-driven attention}
The attention branch normalizes and spikes its input, then projects, normalizes and thresholds the queries, keys and values separately with zero-state gates. For each of $16$ heads of width $64$, causal SDSA~\cite{yao2023sdt,yao2024sdtv2} accumulates key--value co-activations over positions $\ell\leq j\leq t$ and gates them with the current query,
\begin{equation}\small
 \begin{gathered}
 \mathbf m_t=\sum_{\ell\leq j\leq t}\mathbf k_j\odot\mathbf v_j,
 \quad \mathbf h_t=\mathbf q_t\odot\ind[\mathbf m_t\geq1].
 \end{gathered}
 \label{eq:temporal_sdsa}
\end{equation}
A query spike passes when a key and value spike together at some position in this prefix. A linear map projects the concatenated heads to width $192$.

\vspace{1mm}\runin{Causal membrane recurrence in the FFN}
The FFN makes one neural update per context position ($T=1$), passing its membrane state along $\ell,\ldots,t$. It projects the normalized input to $2048$ currents $I_t$ in $16$ groups of $128$ channels. Each group learns a membrane time constant $\tau_g$, with retention $\alpha_g=1-\tau_g^{-1}=0.05+0.9\operatorname{sigmoid}(r_g)$. Initializing $r_g=0$ gives $\tau_g=2$. For one neuron, we adapt \cref{eq:lif_prelim} using input increment $\tfrac12 I_t$ and subtractive reset:
\begin{equation}\small
 u_t=\alpha_g v_{t-1}+\tfrac12 I_t,
 ~s_t=\ind[u_t\geq V_{\mathrm{th}}],~v_t=u_t-V_{\mathrm{th}}\sg(s_t).
 \label{eq:real_history_lif}
\end{equation}
We use $V_{\mathrm{th}}=0.25$. Here $\sg$ denotes stop-gradient, which preserves $s_t$ in the forward pass and sets its derivative to zero in the reset branch. A linear map projects the FFN spikes back to width $192$.

Each predictor call starts from $v_{\ell-1}=0$. Expanding the recurrence gives the pre-fire potential
\begin{equation}
 u_t=\tfrac12\sum_{j=\ell}^{t}\alpha_g^{t-j}I_j
 -V_{\mathrm{th}}\sum_{j=\ell}^{t-1}\alpha_g^{t-j}\sg(s_j).
 \label{eq:causal_membrane_history}
\end{equation}
The first sum integrates the currents, while the second subtracts earlier reset contributions. Since $j\leq t$, this history is causal and limited to at most $L$ context positions. During planning, each candidate rebuilds this history from the observed and predicted latents in its own context window. CMP's output projection then produces the successor latents in \cref{eq:ead_chain}.

\section{Experiments}
\label{sec:experiments}
\edef\experimenttopnumber{\number\value{topnumber}}
\setcounter{topnumber}{1}
\begin{figure*}[t]
  \centering
  \includegraphics[width=\textwidth]{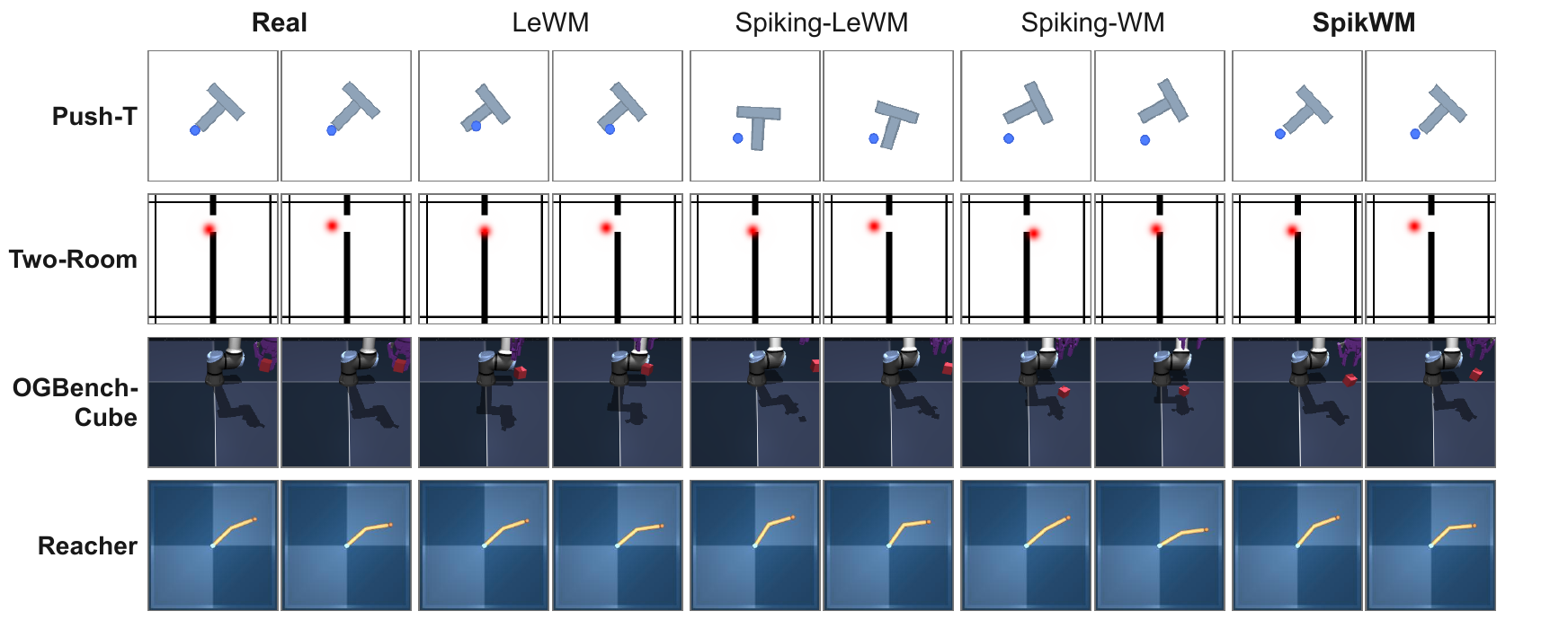}
  \vspace{-6mm}
  \caption{\textbf{Open-loop forecasts on four tasks.} Frame pairs show steps 15/20 (Push-T), 30/35 (Two-Room), 20/35 (OGBench-Cube), and 25/30 (Reacher). We render state readouts of the latent forecasts under shared actions~\cite{maes2026lewm}.}
  \label{fig:decoded_rollouts}
\end{figure*}

\subsection{Benchmarks and Settings}
\runin{Tasks and data}
We use LeWM's offline datasets~\cite{maes2026lewm} for Push-T manipulation~\cite{dinowm2025}, Two-Room navigation~\cite{sobal2025pldm}, Reacher joint-configuration control~\cite{tassa2018deepmindcontrolsuite}, and OGBench-Cube placement~\cite{park2025ogbench}. Training uses four-frame windows sampled five environment steps apart, with each action block containing the intervening commands.

\vspace{1mm}\runin{Planning protocol}
We evaluate each SNN on $50$ start--goal pairs per task at each of three seeds ($42$, $2026$, $2027$). Following LeWM's released protocol, the goal is $25$ commands after the start and the agent has $50$ commands to reach it. All three SNNs use CEM with a five-block horizon and squared terminal latent distance. We implement \spikingbaseline{} with single-update LIF activations. For Spiking-WM~\cite{sun2025spiking}, we reuse \spikingbaseline{}'s visual encoder and add a spiking projection. We widen its DSSM's deterministic state and hidden layers to $1024$ to align parameter scales, using five neural updates per transition. Both adaptations use next-embedding prediction and SIGReg. All local analyses use the final (epoch-ten) checkpoint for each model. Published results retain their source protocols.

\begin{figure}[t]
\centering
\includegraphics[width=\columnwidth]{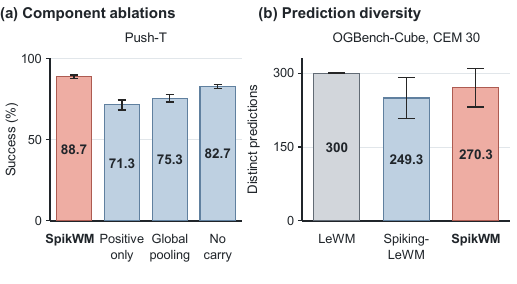}
\vspace{-8mm}
\caption{\textbf{Component ablations and prediction diversity.} (a) Push-T success. (b) Distinct Cube forecasts under $300$ shared action blocks. Bars show mean $\pm$ sample SD across three evaluation seeds (a) and $12$ starts (b).}
\label{fig:pusht_ablation}
\label{fig:component_prediction_analysis}
\end{figure}

\begin{table}[t]
\centering
\caption{\textbf{Push-T success, planning time \& arithmetic energy.} Success means follow \cref{tab:planning_results}. Times are means on one RTX 4090 D. Counts use one common initial-plan request.}
\label{tab:matched_efficiency}
\vspace{-2mm}
\footnotesize
\setlength{\tabcolsep}{2pt}
\renewcommand{\arraystretch}{1.05}
\begin{tabularx}{\linewidth}{@{}l*{5}{>{\raggedleft\arraybackslash}X}@{}}
\toprule
Model & \shortstack{Success\\(\%) $\uparrow$} & \shortstack{FLOPs\\(G)} & \shortstack{SOPs\\(G)} & \shortstack{Energy\\(J) $\downarrow$} & \shortstack{Time\\(s) $\downarrow$}\\
\midrule
LeWM & \textbf{90.40} & 2747.75 & \textbf{0.00} & 6.320 & \textbf{1.53}\\
\spikingbaseline{} & 53.33 & 1771.70 & 65.82 & 4.134 & 2.21\\
Spiking-WM & 37.33 & 2608.01 & 53.97 & 6.047 & 3.82\\
\rowcolor{methodgray}
\textbf{\method{}} & 88.67 & \textbf{600.15} & 178.38 & \textbf{1.541} & 1.97\\
\bottomrule
\end{tabularx}
\end{table}

\begin{figure*}[!t]
  \centering
  \includegraphics[page=1,width=\textwidth]{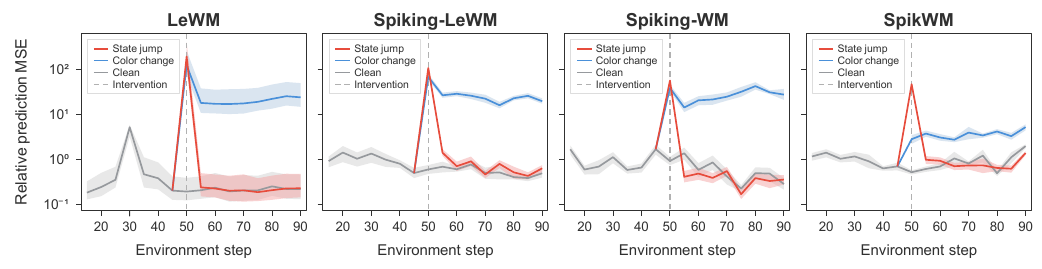}
  \vspace{-6mm}\caption{\textbf{Perturbation responses in Two-Room.} At step 50, we recolor or displace the agent. We divide next-embedding MSE by each model's pre-intervention mean. Bands show 95\% pointwise bootstrap intervals over 16 paired starts.}
  \label{fig:tworoom_surprise}
\end{figure*}

\begin{table}[t]
\centering
\caption{\textbf{Physical probing on OGBench-Cube.} Normalized MSE (SD in subscripts) and mean Pearson $r$.}
\label{tab:physical_probe_cube}
\small
\setlength{\tabcolsep}{2.4pt}
\renewcommand{\arraystretch}{1.02}
\begin{tabularx}{\columnwidth}{l>{\centering\arraybackslash}Xr>{\centering\arraybackslash}Xr}
\toprule
 & \multicolumn{2}{c}{Linear} & \multicolumn{2}{c}{MLP}\\
\cmidrule(lr){2-3}\cmidrule(lr){4-5}
Model & MSE $\downarrow$ & $r\uparrow$ & MSE $\downarrow$ & $r\uparrow$\\
\midrule
\multicolumn{5}{l}{\textit{Gripper position}}\\
LeWM & $0.855_{\pm1.019}$ & $0.633$ & $0.742_{\pm0.805}$ & $0.515$\\
\spikingbaseline{} & $1.007_{\pm1.625}$ & $0.604$ & $0.557_{\pm0.720}$ & $0.666$\\
\rowcolor{methodgray}
\textbf{\method{}} & $\mathbf{0.309}_{\pm0.395}$ & $\mathbf{0.847}$ & $\mathbf{0.394}_{\pm0.544}$ & $\mathbf{0.789}$\\
\midrule
\multicolumn{5}{l}{\textit{Cube position}}\\
LeWM & $\mathbf{0.560}_{\pm1.051}$ & $\mathbf{0.720}$ & $0.655_{\pm0.682}$ & $0.622$\\
\spikingbaseline{} & $0.659_{\pm0.955}$ & $0.685$ & $\mathbf{0.579}_{\pm0.679}$ & $\mathbf{0.672}$\\
\rowcolor{methodgray}
\textbf{\method{}} & $0.620_{\pm0.927}$ & $0.683$ & $0.594_{\pm0.719}$ & $0.658$\\
\midrule
\multicolumn{5}{l}{\textit{Cube yaw}}\\
LeWM & $4.144_{\pm4.171}$ & $\mathbf{0.086}$ & $\mathbf{1.427}_{\pm1.018}$ & $\mathbf{-0.063}$\\
\spikingbaseline{} & $\mathbf{3.362}_{\pm3.619}$ & $-0.172$ & $1.655_{\pm1.001}$ & $-0.152$\\
\rowcolor{methodgray}
\textbf{\method{}} & $3.776_{\pm4.029}$ & $-0.063$ & $1.577_{\pm1.087}$ & $-0.100$\\
\bottomrule
\end{tabularx}
\end{table}

\subsection{Goal-Conditioned Planning}
\runin{Spiking baselines}
SpikWM achieves mean success rates of $88.67\%$ on Push-T, $97.33\%$ on Two-Room, $99.33\%$ on Reacher, and $86.67\%$ on OGBench-Cube. Sample standard deviations range from $1.15\%$ to $3.06\%$ (\cref{tab:planning_results}). It outperforms both spiking baselines on all four tasks. On Push-T, SpikWM uses $16.81$M deployed parameters and reaches $88.67\%$, versus \spikingbaseline{}'s $53.33\%$ with $18.03$M. Compared with Spiking-WM, it uses one neural update per latent position rather than five.

\runin{ANN references}
SpikWM's Push-T mean of $88.67\%$ is below LeWM's reported $90.4\%$. Its means exceed the cited LeWM values on the other three tasks. On Two-Room, $97.33\%$ is below DINO-WM's $99.6\%$ and Qantara's $100\%$. On Reacher, $99.33\%$ exceeds the highest listed ANN mean, TD-JEPA's $97.0\%$. On OGBench-Cube, $86.67\%$ exceeds $\Sigma$-LeWM's $82.6\%$, while Qantara leads at $93.7\%$. These literature comparisons retain their source protocols.

\runin{Appearance and state}
On Reacher and Cube, recoloring increases prediction error less for the spiking models than for LeWM, relative to each model's clean replay. On Reacher, we change the arm color at step $25$ while replaying the same actions. Hard thresholds can suppress color changes that leave firing decisions unchanged. A separate Cube rollout tests how accurately the models predict the gripper's position relative to the cube. The mean error on this rollout is $8.43$ cm for SpikWM and $11.37$ cm for LeWM.

\vspace{-4pt}
\subsection{Ablations}
\vspace{-4pt}
\label{sec:component_experiments}
\Cref{fig:pusht_ablation}a reports mean Push-T success. SpikWM achieves $88.67\%$ with a sample SD of $1.15\%$. The Positive only variant retains BAE's positive stream and achieves $71.33\%$ (SD $3.06\%$). This is the largest drop, suggesting that both polarities help distinguish pushes. Global pooling replaces the four regional averages with a global average per channel, yielding $75.33\%$ (SD $2.31\%$). This decrease is consistent with the need to retain pusher--block geometry. The No carry variant resets the FFN membrane at each position while retaining causal attention, yielding $82.67\%$ (SD $1.15\%$). Each position can still attend to earlier spikes. The six-point drop suggests that carrying subthreshold activity across positions also helps the planner choose a sequence of pushes that moves the block toward its goal pose.

\vspace{-4pt}
\subsection{Predictive Behavior}
\vspace{-3pt}
\label{sec:diagnostic_experiments}
\runin{Action-conditioned prediction diversity}
For \cref{fig:pusht_ablation}b, each model receives the same $300$ action blocks at $12$ Cube starting contexts. We count distinct predicted latent vectors after the first action block. SpikWM retains $270.33\pm39.10$ distinct predictions, compared with $249.33\pm41.42$ for \spikingbaseline{} and $300$ for LeWM. SpikWM therefore distinguishes more action candidates than \spikingbaseline{}. The count measures action resolution but does not indicate whether each predicted future is physically accurate.

\runin{Open-loop forecasts}
\Cref{fig:decoded_rollouts} compares open-loop forecasts from the same observed context and recorded actions. In Push-T, SpikWM keeps the pusher near the T-block's stem, while both spiking baselines predict different block angles. In Two-Room, it follows the agent into the left room, while Spiking-WM leaves it near the doorway. On Cube, it better preserves the gripper-to-cube relation, though the cube remains slightly lower than recorded. On Reacher, it follows the forearm's turn toward horizontal, with LeWM giving the closest joint angles.

\runin{Physical latent probing}
We freeze each world model and fit linear and MLP readouts to recover physical state from its embeddings. On Cube, SpikWM has the lowest normalized gripper-position MSE with both readouts: $0.309$ linearly and $0.394$ with the MLP (\cref{tab:physical_probe_cube}). The linear result shows that the positional advantage is accessible without a nonlinear decoder. End-effector position provides a useful cue for approaching and placing the cube. The advantage is limited to some properties: \spikingbaseline{} has slightly lower MLP cube-position error, and yaw correlations remain weak across the three models.

\runin{Visual and state changes}
We replay the same actions from $16$ paired Two-Room starts. At step $50$, we either recolor the agent in place or move it to a new position (\cref{fig:tworoom_surprise}). Relative to the clean replay at that step, SpikWM's next-embedding error rises $5.40$-fold after recoloring and $89.88$-fold after displacement. Across sampled steps 55--90, its mean recolored-to-clean error ratio is $4.35$, compared with $44.81$--$92.12$ for the baselines. Thus the predictor registers the change in position while remaining less sensitive to the agent's new color over subsequent steps.

\subsection{Planning Efficiency}
Each model plans from one observation and a goal image on the same RTX 4090 D, scoring $300$ candidates per CEM iteration for $30$ iterations (\cref{tab:matched_efficiency}). SpikWM uses $600.15$ GFLOPs and $178.38$ billion synaptic operations per plan. Its estimated arithmetic energy is $1.541$~J, $75.6\%$ below LeWM and $62.7\%$ below \spikingbaseline{}. Across $50$ requests, SpikWM averages $1.97$~s per plan, compared with $2.21$~s for \spikingbaseline{}, $3.82$~s for Spiking-WM, and $1.53$~s for LeWM. Timing includes image encoding, rollouts, scoring, and CEM updates.

\setcounter{topnumber}{\experimenttopnumber}

\FloatBarrier
\section{Conclusion}
We introduced \method{}, an end-to-end JEPA-style spiking world model for goal-conditioned visual planning. Regional pooling retains spatial cues, two spike streams encode complementary action responses, and membrane carry integrates evidence across latent positions. This design uses one simulation step per image and one neural update per latent position. Mean planning success reaches $88.67\%$ on Push-T, $97.33\%$ on Two-Room, $99.33\%$ on Reacher, and $86.67\%$ on OGBench-Cube. Both spiking baselines score lower on all four tasks, and each component ablation reduces Push-T success under the same planner.

{\small
\bibliographystyle{ieeenat_firstcite}
\bibliography{main}}

@article{sun2025spiking,
  author  = {Yinqian Sun and Feifei Zhao and Mingyang Lyu and Yi Zeng},
  title   = {Spiking world model with multicompartment neurons for model-based reinforcement learning},
  journal = {Proceedings of the National Academy of Sciences},
  year    = {2025},
  volume  = {122},
  number  = {50},
  pages   = {e2513319122},
  doi     = {10.1073/pnas.2513319122},
  url     = {https://doi.org/10.1073/pnas.2513319122}
}

@inproceedings{dong2026prednext,
  author = {Yiting Dong and Jianhao Ding and Zijie Xu and Tong Bu and Zhaofei Yu and Tiejun Huang},
  title = {{PredNext}: Explicit Cross-View Temporal Prediction for Unsupervised Learning in Spiking Neural Networks},
  booktitle = {International Conference on Learning Representations},
  year = {2026},
  url = {https://openreview.net/forum?id=LjugJFmItY}
}

@article{rueckert2016planning,
  author  = {Elmar Rueckert and David Kappel and Daniel Tanneberg and Dejan Pecevski and Jan Peters},
  title   = {Recurrent Spiking Networks Solve Planning Tasks},
  journal = {Scientific Reports},
  year    = {2016},
  volume  = {6},
  pages   = {21142},
  doi     = {10.1038/srep21142}
}

@inproceedings{tanneberg2016humanoids,
  author    = {Daniel Tanneberg and Alexandros Paraschos and Jan Peters and Elmar Rueckert},
  title     = {Deep spiking networks for model-based planning in humanoids},
  booktitle = {IEEE-RAS International Conference on Humanoid Robots},
  year      = {2016},
  pages     = {656--661},
  doi       = {10.1109/HUMANOIDS.2016.7803344}
}

@article{basanisi2020planning,
  author  = {Ruggero Basanisi and Andrea Brovelli and Emilio Cartoni and Gianluca Baldassarre},
  title   = {A generative spiking neural-network model of goal-directed behaviour and one-step planning},
  journal = {PLOS Computational Biology},
  year    = {2020},
  volume  = {16},
  number  = {12},
  pages   = {e1007579},
  doi     = {10.1371/journal.pcbi.1007579}
}

@article{capone2024dreaming,
  author  = {Cristiano Capone and Pier Stanislao Paolucci},
  title   = {Towards biologically plausible model-based reinforcement learning in recurrent spiking networks by dreaming new experiences},
  journal = {Scientific Reports},
  year    = {2024},
  volume  = {14},
  pages   = {14656},
  doi     = {10.1038/s41598-024-65631-y}
}

@article{blakowski2025dreaming,
  author  = {Ingo Blakowski and Dmitrii Zendrikov and Giacomo Indiveri and Cristiano Capone},
  title   = {Neuromorphic dreaming as a pathway to efficient learning in artificial agents},
  journal = {Neuromorphic Computing and Engineering},
  year    = {2025},
  volume  = {5},
  pages   = {044005},
  doi     = {10.1088/2634-4386/ae0a77}
}

@mastersthesis{buccapatnam2026spiking,
  author = {Buccapatnam Tirumala, Venkata Saketh Rohit},
  title  = {Spiking World Models for Sample-Efficient Continuous Control: Integrating Population Coding and Model-Based Dreaming},
  school = {Ohio University},
  year   = {2026},
  url    = {https://rave.ohiolink.edu/etdc/view?acc_num=ohiou1776873090921592}
}

@inproceedings{fang2021plif,
  author    = {Fang, Wei and Yu, Zhaofei and Chen, Yanqi and Masquelier, Timoth\'ee and Huang, Tiejun and Tian, Yonghong},
  title     = {Incorporating Learnable Membrane Time Constant To Enhance Learning of Spiking Neural Networks},
  booktitle = {Proceedings of the IEEE/CVF International Conference on Computer Vision},
  year      = {2021},
  month     = {October},
  pages     = {2661--2671},
  url       = {https://openaccess.thecvf.com/content/ICCV2021/html/Fang_Incorporating_Learnable_Membrane_Time_Constant_To_Enhance_Learning_of_Spiking_ICCV_2021_paper.html}
}

@inproceedings{dinowm2025,
  author    = {Zhou, Gaoyue and Pan, Hengkai and LeCun, Yann and Pinto, Lerrel},
  editor    = {Singh, Aarti and Fazel, Maryam and Hsu, Daniel and Lacoste-Julien, Simon and Berkenkamp, Felix and Maharaj, Tegan and Wagstaff, Kiri and Zhu, Jerry},
  title     = {{DINO}-{WM}: World Models on Pre-trained Visual Features enable Zero-shot Planning},
  booktitle = {Proceedings of the 42nd International Conference on Machine Learning},
  year      = {2025},
  month     = {13--19 Jul},
  volume    = {267},
  pages     = {79115--79135},
  publisher = {PMLR},
  series    = {Proceedings of Machine Learning Research},
  url       = {https://proceedings.mlr.press/v267/zhou25t.html}
}

@misc{spikeworld2026,
  author        = {Ziqiao Yu},
  title         = {{SpikeWorld}: Fast-State Adaptation for Frozen Spiking World Models},
  year          = {2026},
  howpublished  = {arXiv preprint arXiv:2608.07712},
  url           = {https://arxiv.org/abs/2608.07712},
  eprint        = {2608.07712},
  archiveprefix = {arXiv},
  primaryclass  = {cs.CV}
}

@misc{tassa2018deepmindcontrolsuite,
  title = {DeepMind Control Suite},
  author = {Yuval Tassa and Yotam Doron and Alistair Muldal and Tom Erez and Yazhe Li and Diego de Las Casas and David Budden and Abbas Abdolmaleki and Josh Merel and Andrew Lefrancq and Timothy Lillicrap and Martin Riedmiller},
  year = {2018},
  eprint = {1801.00690},
  archiveprefix = {arXiv},
  primaryclass = {cs.AI},
  url = {https://arxiv.org/abs/1801.00690},
  howpublished = {arXiv preprint arXiv:1801.00690}
}

@inproceedings{hafner2020dreamcontrollearningbehaviors,
  title = {Dream to Control: Learning Behaviors by Latent Imagination},
  author = {Danijar Hafner and Timothy Lillicrap and Jimmy Ba and Mohammad Norouzi},
  year = {2020},
  eprint = {1912.01603},
  archiveprefix = {arXiv},
  primaryclass = {cs.LG},
  url = {https://openreview.net/forum?id=S1lOTC4tDS},
  booktitle = {International Conference on Learning Representations}
}

@inproceedings{hafner2019planet,
  title      = {Learning Latent Dynamics for Planning from Pixels},
  author     = {Hafner, Danijar and Lillicrap, Timothy and Fischer, Ian and Villegas, Ruben and Ha, David and Lee, Honglak and Davidson, James},
  booktitle  = {Proceedings of the 36th International Conference on Machine Learning},
  pages      = {2555--2565},
  year       = {2019},
  volume     = {97},
  series     = {Proceedings of Machine Learning Research},
  publisher  = {PMLR},
  url        = {https://proceedings.mlr.press/v97/hafner19a.html}
}

@inproceedings{nguyen2021tpc,
  title      = {Temporal Predictive Coding For Model-Based Planning In Latent Space},
  author     = {Nguyen, Tung D and Shu, Rui and Pham, Tuan and Bui, Hung and Ermon, Stefano},
  booktitle  = {Proceedings of the 38th International Conference on Machine Learning},
  pages      = {8130--8139},
  year       = {2021},
  volume     = {139},
  series     = {Proceedings of Machine Learning Research},
  publisher  = {PMLR},
  url        = {https://proceedings.mlr.press/v139/nguyen21h.html}
}

@misc{maes2026lewm,
  title = {{LeWorldModel}: Stable End-to-End Joint-Embedding Predictive Architecture from Pixels},
  author = {Lucas Maes and Quentin Le Lidec and Damien Scieur and Yann LeCun and Randall Balestriero},
  year = {2026},
  eprint = {2603.19312},
  archiveprefix = {arXiv},
  primaryclass = {cs.LG},
  url = {https://arxiv.org/abs/2603.19312},
  howpublished = {arXiv preprint arXiv:2603.19312}
}

@inproceedings{zhou2023spikformer,
      title={{Spikformer}: When Spiking Neural Network Meets Transformer},
      author={Zhaokun Zhou and Yuesheng Zhu and Chao He and Yaowei Wang and Shuicheng Yan and Yonghong Tian and Li Yuan},
      year={2023},
      booktitle={International Conference on Learning Representations},
      eprint={2209.15425},
      archivePrefix={arXiv},
      primaryClass={cs.NE},
      url={https://arxiv.org/abs/2209.15425},
}

@inproceedings{yao2023sdt,
  title = {Spike-driven Transformer},
  author = {Man Yao and Jiakui Hu and Zhaokun Zhou and Li Yuan and Yonghong Tian and Bo Xu and Guoqi Li},
  year = {2023},
  booktitle = {Advances in Neural Information Processing Systems},
  eprint = {2307.01694},
  archiveprefix = {arXiv},
  primaryclass = {cs.NE},
  url = {https://proceedings.neurips.cc/paper_files/paper/2023/hash/ca0f5358dbadda74b3049711887e9ead-Abstract-Conference.html},
  volume = {36},
  pages = {64043--64058},
  doi = {10.52202/075280-2798}
}

@inproceedings{yao2024sdtv2,
      title={Spike-driven Transformer V2: Meta Spiking Neural Network Architecture Inspiring the Design of Next-generation Neuromorphic Chips},
      author={Man Yao and Jiakui Hu and Tianxiang Hu and Yifan Xu and Zhaokun Zhou and Yonghong Tian and Bo Xu and Guoqi Li},
      year={2024},
      booktitle={International Conference on Learning Representations},
      eprint={2404.03663},
      archivePrefix={arXiv},
      primaryClass={cs.NE},
      url={https://arxiv.org/abs/2404.03663},
}

@article{yao2025sdtv3,
  author = {Man Yao and Xuerui Qiu and Tianxiang Hu and Jiakui Hu and Yuhong Chou and Keyu Tian and Jianxing Liao and Luziwei Leng and Bo Xu and Guoqi Li},
  title = {Scaling Spike-Driven Transformer With Efficient Spike Firing Approximation Training},
  journal = {IEEE Transactions on Pattern Analysis and Machine Intelligence},
  year = {2025},
  volume = {47},
  number = {4},
  pages = {2973--2990},
  doi = {10.1109/tpami.2025.3530246},
  url = {https://doi.org/10.1109/tpami.2025.3530246},
}

@misc{ye2026fire,
  author={Ye, Shuhan and Qian, Yuanbin and Yu, Yi and Wang, Chong and Xie, Yuqi and Xu, Jiazhen and Wang, Kun and Jiang, Xudong},
  title={Fire on Motion: Optimizing Video Pass-bands for Efficient Spiking Action Recognition},
  year={2026},
  howpublished={arXiv preprint arXiv:2601.22675},
  url={https://arxiv.org/abs/2601.22675}
}

@InProceedings{zou2025spikevideoformer,
  title = 	 {{S}pike{V}ideo{F}ormer: An Efficient Spike-Driven Video Transformer with Hamming Attention and $\mathcal{O}(T)$ Complexity},
  author =       {Zou, Shihao and Li, Qingfeng and Ji, Wei and Li, Jingjing and Yang, Yongkui and Li, Guoqi and Dong, Chao},
  booktitle = 	 {Proceedings of the 42nd International Conference on Machine Learning},
  pages = 	 {80805--80821},
  year = 	 {2025},
  editor = 	 {Singh, Aarti and Fazel, Maryam and Hsu, Daniel and Lacoste-Julien, Simon and Berkenkamp, Felix and Maharaj, Tegan and Wagstaff, Kiri and Zhu, Jerry},
  volume = 	 {267},
  series = 	 {Proceedings of Machine Learning Research},
  month = 	 {13--19 Jul},
  publisher =    {PMLR},
  url = 	 {https://proceedings.mlr.press/v267/zou25b.html}
}

@misc{assran2025vjepa2,
  title = {{V-JEPA 2}: Self-Supervised Video Models Enable Understanding, Prediction and Planning},
  author = {Mido Assran and Adrien Bardes and David Fan and Quentin Garrido and Russell Howes and Mojtaba Komeili and Matthew Muckley and Ammar Rizvi and Claire Roberts and Koustuv Sinha and Artem Zholus and Sergio Arnaud and Abha Gejji and Ada Martin and Francois Robert Hogan and Daniel Dugas and Piotr Bojanowski and Vasil Khalidov and Patrick Labatut and Francisco Massa and Marc Szafraniec and Kapil Krishnakumar and Yong Li and Xiaodong Ma and Sarath Chandar and Franziska Meier and Yann LeCun and Michael Rabbat and Nicolas Ballas},
  year = {2025},
  eprint = {2506.09985},
  archiveprefix = {arXiv},
  primaryclass = {cs.AI},
  url = {https://arxiv.org/abs/2506.09985},
  howpublished = {arXiv preprint arXiv:2506.09985}
}

@inproceedings{assran2023ijepa,
  author    = {Assran, Mahmoud and Duval, Quentin and Misra, Ishan and Bojanowski, Piotr and Vincent, Pascal and Rabbat, Michael and LeCun, Yann and Ballas, Nicolas},
  title     = {Self-Supervised Learning From Images With a Joint-Embedding Predictive Architecture},
  booktitle = {Proceedings of the IEEE/CVF Conference on Computer Vision and Pattern Recognition},
  year      = {2023},
  pages     = {15619--15629},
  url       = {https://openaccess.thecvf.com/content/CVPR2023/html/Assran_Self-Supervised_Learning_From_Images_With_a_Joint-Embedding_Predictive_Architecture_CVPR_2023_paper.html}
}

@misc{zhang2026sgjepa,
  title = {Scalable and Efficient Joint Spiking Embedding Predictive Architecture for Large-Scale Dynamic Graphs},
  author = {Huizhe Zhang and Yuchang Zhu and Huazhen Zhong and Liang Chen and Zibin Zheng},
  year = {2026},
  eprint = {2607.18412},
  archiveprefix = {arXiv},
  primaryclass = {cs.LG},
  url = {https://arxiv.org/abs/2607.18412},
  howpublished = {arXiv preprint arXiv:2607.18412}
}

@article{fan2025msf,
  title = {A multisynaptic spiking neuron for simultaneously encoding spatiotemporal dynamics},
  author = {Liangwei Fan and Hui Shen and Xiangkai Lian and Yulin Li and Man Yao and Guoqi Li and Dewen Hu},
  journal = {Nature Communications},
  year = {2025},
  doi = {10.1038/s41467-025-62251-6},
  url = {https://doi.org/10.1038/s41467-025-62251-6},
  volume = {16},
  number = {1},
  pages = {7155}
}

@article{zhang2026hmt,
  title = {Generalized Threshold Optimization with Harmony Multi-Threshold Neurons for Accurate ANN-to-SNN Conversion},
  author = {Wenhan Zhang and Zihan Huang and Tong Bu and Tiejun Huang and Zhaofei Yu},
  journal = {Proceedings of the AAAI Conference on Artificial Intelligence},
  year = {2026},
  doi = {10.1609/aaai.v40i3.37206},
  url = {https://doi.org/10.1609/aaai.v40i3.37206},
  volume = {40},
  number = {3},
  pages = {2227--2235}
}

@inproceedings{sobal2025pldm,
  title = {Learning from Reward-Free Offline Data: A Case for Planning with Latent Dynamics Models},
  author = {Sobal, Uladzislau and Zhang, Wancong and Cho, Kyunghyun and Balestriero, Randall and Rudner, Tim G. J. and LeCun, Yann},
  booktitle = {Advances in Neural Information Processing Systems},
  volume = {38},
  year = {2025},
  url = {https://proceedings.neurips.cc/paper_files/paper/2025/hash/3e7cf447f21cd11c846463affefce665-Abstract-Conference.html},
  pages = {49126--49162},
  doi = {10.52202/085713-1465}
}

@misc{yu2024svformerdirecttrainingspiking,
  title = {{SVFormer}: A Direct Training Spiking Transformer for Efficient Video Action Recognition},
  author = {Liutao Yu and Liwei Huang and Chenlin Zhou and Han Zhang and Zhengyu Ma and Huihui Zhou and Yonghong Tian},
  year = {2024},
  eprint = {2406.15034},
  archiveprefix = {arXiv},
  primaryclass = {cs.CV},
  url = {https://arxiv.org/abs/2406.15034},
  howpublished = {arXiv preprint arXiv:2406.15034}
}

@misc{gleyo2026spikingjepa,
  title={Bio-plausible statistical regularizer for spiking {JEPA}},
  author={Alexis Gleyo and Laurent Perrinet}, year={2026},
  howpublished={SNUFA 2026 workshop abstract},
  url={https://snufa.net/2026/abstracts/alexis-gleyo-bio-plausible.html}
}

@inproceedings{park2025ogbench,
  author={Park, Seohong and Frans, Kevin and Eysenbach, Benjamin and Levine, Sergey},
  title={{OGBench}: Benchmarking Offline Goal-Conditioned {RL}},
  booktitle={International Conference on Learning Representations},
  year={2025},
  url={https://proceedings.iclr.cc/paper_files/paper/2025/hash/ecd92623ac899357312aaa8915853699-Abstract-Conference.html}
}

@inproceedings{tang2021popsan,
  author={Tang, Guangzhi and Kumar, Neelesh and Yoo, Raymond and Michmizos, Konstantinos},
  title={Deep Reinforcement Learning with Population-Coded Spiking Neural Network for Continuous Control},
  booktitle={Proceedings of the 2020 Conference on Robot Learning},
  series={Proceedings of Machine Learning Research},
  volume={155},
  pages={2016--2029},
  year={2021},
  url={https://proceedings.mlr.press/v155/tang21a.html}
}

@misc{balestriero2025lejepa,
  author = {Balestriero, Randall and LeCun, Yann},
  title = {{LeJEPA}: Provable and Scalable Self-Supervised Learning Without the Heuristics},
  year = {2025},
  eprint = {2511.08544},
  archiveprefix = {arXiv},
  url = {https://arxiv.org/abs/2511.08544},
  howpublished = {arXiv preprint arXiv:2511.08544}
}

@article{rubinstein1999cem,
  author={Rubinstein, Reuven},
  title={The Cross-Entropy Method for Combinatorial and Continuous Optimization},
  journal={Methodology and Computing in Applied Probability},
  volume={1},
  number={2},
  pages={127--190},
  year={1999},
  doi={10.1023/A:1010091220143},
  url={https://doi.org/10.1023/A:1010091220143}
}

@article{eshraghian2023snn,
  author={Eshraghian, Jason K. and Ward, Max and Neftci, Emre O. and Wang, Xinxin and Lenz, Gregor and Dwivedi, Girish and Bennamoun, Mohammed and Jeong, Doo Seok and Lu, Wei D.},
  title={Training Spiking Neural Networks Using Lessons From Deep Learning},
  journal={Proceedings of the IEEE},
  year={2023},
  volume={111},
  number={9},
  pages={1016--1054},
  doi={10.1109/JPROC.2023.3308088},
  url={https://doi.org/10.1109/JPROC.2023.3308088}
}

@article{neftci2019sg,
  author={Neftci, Emre O. and Mostafa, Hesham and Zenke, Friedemann},
  title={Surrogate Gradient Learning in Spiking Neural Networks: Bringing the Power of Gradient-Based Optimization to Spiking Neural Networks},
  journal={IEEE Signal Processing Magazine},
  year={2019},
  volume={36},
  number={6},
  pages={51--63},
  doi={10.1109/MSP.2019.2931595},
  url={https://doi.org/10.1109/MSP.2019.2931595}
}

@misc{achituve2026adaptivelatentcapacityworld,
  title = {Adaptive Latent Capacity for World Models},
  author = {Idan Achituve and Lior Dikstein and Idit Diamant and Arnon Netzer and Hai Victor Habi},
  year = {2026},
  eprint = {2609.32921},
  archiveprefix = {arXiv},
  primaryclass = {cs.LG},
  url = {https://arxiv.org/abs/2609.32921},
  howpublished = {arXiv preprint arXiv:2609.32921}
}

@misc{gao2026fastlewm,
  author={Yuntian Gao and Xiangyu Xu},
  title={Fast {LeWorldModel}},
  year={2026},
  howpublished={arXiv:2606.26217},
  url={https://arxiv.org/abs/2606.26217}
}

@misc{zu2026sigmalewm,
  author={Xinrui Zu and Michael Przystupa and Zhao Yang and Shujian Yu and Kevin Sebastian Luck},
  title={{$\Sigma$-LeWM}: Learning World Models via {S}-Information for Robot Control},
  year={2026},
  howpublished={arXiv:2611.12732},
  url={https://arxiv.org/abs/2611.12732}
}

@misc{zhao2026subjepa,
  author={Kai Zhao and Dongliang Nie and Yuchen Lin and Zhehan Luo and Yixiao Gu and Deng-Ping Fan and Dan Zeng},
  title={{Sub-JEPA}: Subspace Gaussian Regularization for Stable End-to-End World Models},
  year={2026},
  howpublished={arXiv:2605.09241},
  url={https://arxiv.org/abs/2605.09241}
}

@misc{wang2026anchorwm,
  author={Yanbo Wang and Nan Lin and Wendelin B{\"o}hmer and Javier Alonso-Mora},
  title={Pretrained Features as Teachers, Not States: Compact Latent World Models for Visual Planning},
  year={2026},
  howpublished={arXiv:2610.23984},
  url={https://arxiv.org/abs/2610.23984}
}

@misc{li2026rcaux,
  author={Wenyuan Li and Guang Li and Keisuke Maeda and Takahiro Ogawa and Miki Haseyama},
  title={Predictive but Not Plannable: {RC-aux} for Latent World Models},
  year={2026},
  howpublished={arXiv:2605.07278},
  doi={10.48550/arXiv.2605.07278},
  url={https://arxiv.org/abs/2605.07278v1}
}

@misc{bai2026temporaldistancejepa,
  author={Jiaxin Bai and Jiaxuan Xiong},
  title={Temporal-Distance {JEPA}: Plan-Aware Representation Learning for Latent World Model Predictive Control},
  year={2026},
  howpublished={arXiv:2607.25337},
  doi={10.48550/arXiv.2607.25337},
  url={https://arxiv.org/abs/2607.25337v2}
}

@misc{rakhimov2026qantara,
  author={Ruslan Rakhimov and George Bredis and Yuriy Maksyuta and Daniil Gavrilov},
  title={{Qantara}: Bridge-Flow Training for Multi-Paradigm {JEPA} Inference},
  year={2026},
  howpublished={arXiv:2607.04978},
  doi={10.48550/arXiv.2607.04978},
  url={https://arxiv.org/abs/2607.04978v2}
}

@inproceedings{xu2024feelsnn,
  author = {Mengting Xu and De Ma and HuaJin Tang and Qian Zheng and Gang Pan},
  title = {{FEEL-SNN}: Robust Spiking Neural Networks with Frequency Encoding and Evolutionary Leak Factor},
  booktitle = {Advances in Neural Information Processing Systems},
  volume = {37},
  year = {2024},
  doi = {10.52202/079017-2917},
  pages = {91930--91950},
  url = {https://proceedings.neurips.cc/paper_files/paper/2024/hash/a73474c359ed523e6cd3174ed29a4d56-Abstract-Conference.html}
}

@inproceedings{fang2025highfrequency,
  author = {Yuetong Fang and Deming Zhou and Ziqing Wang and Hongwei Ren and Zecui Zeng and Lusong Li and Shibo Zhou and Renjing Xu},
  title = {Spiking Neural Networks Need High-Frequency Information},
  booktitle = {Advances in Neural Information Processing Systems},
  volume = {38},
  year = {2025},
  doi = {10.52202/085713-3455},
  pages = {114506--114527},
  url = {https://proceedings.neurips.cc/paper_files/paper/2025/hash/956834836f36dd07df7064ff42ca69f2-Abstract-Conference.html}
}
\end{document}